\documentclass[10pt,twocolumn,letterpaper]{article}
\usepackage[left=0.75in,right=0.75in,top=0.75in,bottom=0.85in,columnsep=0.33in]{geometry}
\usepackage{fontspec}
\usepackage{balance}
\usepackage{amsmath}
\usepackage{unicode-math}
\usepackage{graphicx,booktabs,array,xcolor,tikz,caption,titlesec,enumitem,microtype}
\usepackage{float}
\usetikzlibrary{arrows.meta,positioning}
\definecolor{navy}{HTML}{1F3A5F}\definecolor{acc}{HTML}{B5483A}\definecolor{slate}{HTML}{7C8FA6}\definecolor{ink}{HTML}{2B2F36}\definecolor{mist}{HTML}{EEF1F5}
\titleformat{\section}{\normalfont\fontsize{11}{13}\bfseries\centering}{\thesection}{0.8em}{}
\titlespacing*{\section}{0pt}{10pt plus 2pt minus 1pt}{5pt}
\titleformat{\subsection}[runin]{\normalfont\normalsize\bfseries}{}{0pt}{}[\quad]
\titlespacing*{\subsection}{0pt}{5pt plus 1pt}{0pt}
\begin{document}
\twocolumn[{\centering{\fontsize{16}{20}\selectfont\bfseries Closed-Loop Agent Evaluation:\\Layered Measurement and Self-Calibration\par}\vspace{10pt}\begingroup\linespread{1}\selectfont {\large\bfseries Ruifeng Ren*, Gun Ahn}\\[3pt]{\small Massachusetts Institute of Technology, Cambridge, MA, USA\\ rayren@alum.mit.edu, guna23@mit.edu}\endgroup\par\vspace{14pt}}]
\begingroup\leftskip=8pt\rightskip=8pt\centerline{\bfseries\large Abstract}\vspace{3pt}\noindent\small Once an agent is live, teams want to know whether it is doing its job, not just whether it throws errors. Yet most evaluation is one-off: a benchmark run, a look at traces, a single score. Its conclusions neither check themselves nor reach the agent. We propose \textbf{closed-loop evaluation}, an evaluation method for deployed agents. It models evaluation as performance estimation under biased raters and censored outcomes, and it requires every level’s conclusion to trace back to atomic data and to be computed by versioned rules. The method covers four levels of analysis (span, trace, conversation and engagement), measures along a spine of Execution, Substance, Trajectory and Results, treats Conduct as a constraint throughout, and uses Provenance and Adjudication components to certify the evidence and the evaluators. Evaluator agents generated by an Agent Factory carry these components, each with a domain knowledge pack and an adapter harness. Two loops close the evaluation: an inner loop audits the evidence, rules and metrics themselves, and an outer loop uses per-component regression checks to drive agent improvement. We also give a pipeline that runs this process automatically. Using the business-analysis agent SharpInsight as a case and porting the method to public conversation data, we show how problems differ across levels and how evidence-and-rule judgment is more reliable than holistic scoring (ICC 0.86 vs.\ 0.81; κ 0.81 vs.\ 0.63).\par\endgroup\vspace{6pt}
\let\thefootnote\relax\footnotetext{\fontsize{7.5}{9}\selectfont * Corresponding author: rayren@alum.mit.edu}
\section{Introduction}
Enterprises increasingly deploy LLM agents the way they staff a role: the agent works independently, faces customers directly and operates under a specification (its system prompt). Yet it is still evaluated like software, with offline benchmarks, trace inspection or an LLM judge deciding whether an answer is good. These practices share two flaws. First, \textbf{evaluation is open-loop}: once a score is computed the process ends, nobody checks whether it can be trusted, and nothing ensures it changes the agent. Second, \textbf{judgment is a black box}: an LLM returns a score without showing what it consists of, and another model may conclude differently.

Our approach is to review an agent the way an organization reviews an employee. Nobody reviews an employee with one person’s overall score. A review examines attendance records, work product, client feedback and business results against explicit standards, and it calibrates the raters. Personnel research has long found that a rater’s own tendencies can explain more of the variance in ratings than the performance being rated (Scullen, Mount, and Goff 2000), and that rewarding proxies distorts behavior (Kerr 1975; Holmström and Milgrom 1991). Agent evaluation faces the same problems, at a scale that demands automation.

We make four contributions: (1) a \textbf{closed-loop evaluation method} comprising a measurement model, four levels of analysis, seven components and two loops; (2) an architecture that \textbf{separates atomic data acquisition from deterministic analysis}, with data-level audits; (3) an Agent Factory pipeline that makes the loop run \textbf{automatically}; and (4) empirical runs on a real agent and a public dataset showing how the method operates and ports.

\section{The Closed-Loop Evaluation Method}
\begin{figure*}[t]\centering
\resizebox{\linewidth}{!}{\begin{tikzpicture}[x=1pt,y=1pt,font=\sffamily\fontsize{7.6}{9.5}\selectfont,
 bar/.style={draw=#1,line width=.5pt,rounded corners=1.5pt,minimum height=15pt,inner sep=1pt},
 st/.style={draw=navy,fill=mist,line width=.5pt,rounded corners=1.5pt,minimum width=76pt,minimum height=34pt,align=center,inner sep=1pt},
 arr/.style={-{Stealth[length=4pt,width=3pt]},line width=.6pt,color=ink}]
\node[bar=slate,fill=white,minimum width=336pt] (a) at (182,100) {\textbf{Adjudication: can the evaluators and metrics be trusted?}};
\node[st] (e) at (52,70) {\textbf{Execution}\\[1pt]{\color{ink!75}\fontsize{6.6}{8}\selectfont is the agent at work?}};
\node[st] (s) at (138,70) {\textbf{Substance}\\[1pt]{\color{ink!75}\fontsize{6.6}{8}\selectfont is the work good?}};
\node[st] (t) at (226,70) {\textbf{Trajectory}\\[1pt]{\color{ink!75}\fontsize{6.6}{8}\selectfont what users went through}};
\node[st] (r) at (312,70) {\textbf{Results}\\[1pt]{\color{ink!75}\fontsize{6.6}{8}\selectfont did the outcome hold?}};
\draw[arr] (e)--(s); \draw[arr] (s)--(t); \draw[arr] (t)--(r);
\node[bar=slate,fill=white,minimum width=336pt] (c) at (182,41) {\textbf{Conduct: does the agent follow its specification throughout?}};
\node[bar=slate,fill=white,minimum width=336pt] (p) at (182,22) {\textbf{Provenance: who is the traffic, and can the evidence be trusted?}};
\draw[-{Stealth[length=5pt,width=4pt]},line width=.8pt,color=navy] (r.east) .. controls (392,70) and (392,-6) .. (182,-6) .. controls (-10,-6) and (-14,70) .. (e.west);
\node[fill=white,inner sep=1.5pt,text=navy] at (182,-6) {\textbf{Outer loop: find issues → fix the agent → per-component regression → ramp up → re-evaluate}};
\node[draw=ink,line width=.5pt,rounded corners=2pt,minimum width=104pt,minimum height=88pt,align=center,inner sep=2pt] (f) at (449,60) {\textbf{\fontsize{8.4}{10}\selectfont Agent Factory}\\[4pt]spec + knowledge + harness\\[2pt]\textbf{atomic data acquisition}\\[2pt]\textbf{deterministic analysis}\\[4pt]= evaluator agent};
\draw[-{Stealth[length=5pt,width=4pt]},line width=.8pt,color=acc] (a.east) .. controls (370,100) and (375,92) .. (f.north west|-0,92);
\node[text=acc,anchor=south] at (396,105) {\textbf{Inner loop: regenerate}};
\draw[-{Stealth[length=4pt,width=3pt]},line width=.6pt,color=acc,dashed] (f.west|-0,41) -- (c.east);
\end{tikzpicture}}
\caption{The closed-loop evaluation method. The center shows the four-stage spine, with Conduct running across it and Provenance and Adjudication below and above. The red inner loop corrects evaluators; the blue outer loop corrects the agent.}
\vspace{7pt}
\fontsize{8}{10}\selectfont\setlength{\tabcolsep}{4pt}
\begin{tabular}{@{}>{\raggedright\arraybackslash}p{\dimexpr 0.120\linewidth-8pt\relax}>{\raggedright\arraybackslash}p{\dimexpr 0.200\linewidth-8pt\relax}>{\raggedright\arraybackslash}p{\dimexpr 0.360\linewidth-8pt\relax}>{\raggedright\arraybackslash}p{\dimexpr 0.320\linewidth-8pt\relax}@{}}
\toprule
\textbf{Component} & \textbf{Question} & \textbf{Atomic data φ} & \textbf{Deterministic analysis \textit{f}} \\ \midrule
Provenance & Who is the traffic? Can the evidence be trusted? & account and traffic type, cached returns, attacks, evidence coverage & segmentation rules; refuse review if coverage is short \\
Execution & Did the agent run correctly? & step chain, task status, latency, delivery status & per-mode baselines; failure rules \\
Substance & Is the deliverable good? & checklist items & checklist score \textit{S}; item pass rates \\
Trajectory & What did users go through? & interaction recovery events: request replay, context repair, session termination & friction \textit{F}; rule-derived outcome \\
Results & Did the outcome hold? & return, return gap, redo, business events & retention net of redo; censoring flag \\
Conduct & Does it follow its spec? & clause triggered, obligation met & applicability \textit{A}, conditional compliance \textit{C}, invariants \\
Adjudication & Can evaluation be trusted? & paired data across models; rules vs. atomic data & κ, ICC, precision and recall, admission \\
\bottomrule
\end{tabular}
\captionof{table}{Atomic data and deterministic analysis of the seven components. All conclusions come from the rules in the right column.}
\end{figure*}
\textbf{Measurement model and levels.} Analysis runs at four levels: a \textit{span} is one model or tool call, a \textit{trace} is the span tree of one turn, a \textit{conversation} is a session, and an \textit{engagement} is a user’s conversations and outcome events within horizon \textit{T}. Higher levels sit closer to user value, but their evidence is sparser and later. For unit \textit{x}, the output of evaluator \textit{e} is modeled as

\[ y_e(x) = g_e\bigl(\theta(x)\bigr) + b_e(x) + \varepsilon_e \]
\noindent where θ is latent performance, \textit{g}\textsubscript{e} a monotone link and \textit{b}\textsubscript{e} rater bias, and the output is visible only when an observation mask \textit{M}\textsubscript{e}(\textit{x}) = 1. The model yields three requirements. A single rater confounds \textit{b}\textsubscript{e} with θ, so multiple calibrated raters are needed. Pooling segments with different \textit{g}\textsubscript{e} estimates a population that does not exist, so evidence must be certified before scoring. And \textit{M} is not missing at random, since users who fail silently leave first, so unobserved outcomes cannot count as successes.

\textbf{Seven components.} The method sets four spine stages along the path of value creation (Figure~1): \textbf{Execution}, \textbf{Substance}, \textbf{Trajectory} and \textbf{Results}. \textbf{Conduct} runs throughout, since an agent can succeed at every stage and still overstep. \textbf{Provenance} certifies the evidence and \textbf{Adjudication} certifies the evaluators; neither sits on the spine, which would confuse the instrument with the object measured. Table~1 gives each component’s atomic data and analysis, independent of any business domain.

\textbf{Evaluator agents.} Each component is carried by evaluator agents, defined formally as \textit{e} = \ensuremath{\langle}\textit{K}, \textit{H}, φ, \textit{f}, ρ, π, σ, \textit{v}\ensuremath{\rangle}. φ acquires atomic data, by rule or LLM, and each data item is a checkable atomic judgment with no score; \textit{f} is the deterministic analysis, versioned rules that compute results from the data, so the same input always yields the same conclusion. \textit{K} is a \textbf{domain knowledge pack} (latency baselines, behavior lexicons, policy clauses); \textit{H} is an \textbf{adapter harness} that connects to logs and traces, calls tools and normalizes data (OpenTelemetry 2026); ρ holds the data definitions; π is a context projection exposing only the needed evidence; σ is an output schema with evidence pointers; and \textit{v} is a version. Results are cached under \textit{k} = hash(\textit{v} \ensuremath{\|} hash(π(\textit{x}))), so a new turn costs one incremental computation and a definition change invalidates only what it affects. Acquisition follows a cascade: rules first, then a small model when uncertain, then a cross-family jury, which reduces self-preference (Panickssery, Bowman, and Feng 2024; Verga et al. 2024).

Two typical analyses follow. The Substance checklist score is

\[ S(x) = D(x)\cdot\frac{1}{|A(x)|}\sum_{j\in A(x)} a_j(x) \]
\noindent where \textit{D} is rule-derived delivery status and \textit{A}(\textit{x}) the admitted, applicable items. Trajectory friction is \textit{F}(\textit{x}) = \ensuremath{\sum}\textsubscript{s} \textit{a}\textsubscript{s}(\textit{x}) over admitted interaction recovery events, and the outcome follows by rule: \textbf{blocked} on abandonment or frustration; otherwise \textbf{struggled} if \textit{F} \ensuremath{\geq} 1; \textbf{censored} if single-turn with none of these; else \textbf{smooth}.

\textbf{Clauses, censoring and admission.} A specification compiles into clauses \textit{r} = (φ\textsubscript{t}, ψ) of a trigger and an obligation. Rules detect triggers and LLMs only acquire the atomic datum of whether the obligation is met, with evidence, yielding applicability \textit{A}(\textit{r}) = P(φ\textsubscript{t}) and conditional compliance \textit{C}(\textit{r}) = P(ψ | φ\textsubscript{t}). For unseen outcomes we do not impute satisfaction: the Results stage supplies delayed labels (return, gap, and redo, defined by character-bigram Jaccard \ensuremath{\geq} 0.6 between successive requests), and sessions with no return within the horizon are marked censored. Audits work at the data level: replacing \textit{y}\textsubscript{e} in the measurement model with data item \textit{a}\textsubscript{j,e}, each data item must pass cross-model agreement to enter \textit{f}, and metrics must also pass criterion validity, discriminant validity and a Goodhart test (Goodhart 1975), under which a metric must not improve when a lower stage degrades.

\textbf{Two loops.} The \textbf{inner loop} acts on evaluation itself: when a test fails, the data definition, rule, evidence scope or metric changes, the version changes, results are recomputed, and failing data items leave the analysis. The \textbf{outer loop} acts on the agent: findings become fixes, and a new version must pass regression checks on every component, with zero tolerance for invariants and no significant decline elsewhere, before ramping up, and new traffic enters the next round. Components are reported side by side, not as a weighted composite, to avoid multitask distortion.

\section{Data Analysis and Breakdown}
\begin{table*}[t]
\centering
\renewcommand{\thetable}{2}
\fontsize{8}{10}\selectfont\setlength{\tabcolsep}{4pt}
\begin{tabular}{@{}>{\raggedright\arraybackslash}p{\dimexpr 0.110\linewidth-8pt\relax}>{\raggedright\arraybackslash}p{\dimexpr 0.240\linewidth-8pt\relax}>{\raggedright\arraybackslash}p{\dimexpr 0.230\linewidth-8pt\relax}>{\raggedright\arraybackslash}p{\dimexpr 0.250\linewidth-8pt\relax}>{\raggedright\arraybackslash}p{\dimexpr 0.170\linewidth-8pt\relax}@{}}
\toprule
\textbf{Level} & \textbf{Atomic data} & \textbf{Deterministic analysis} & \textbf{Case performance} & \textbf{Only this level answers} \\ \midrule
Span & step type, order, omission; task progress & chain completeness; stage of abnormal stop & no thinking step: 15.7\% fail (0.0\% with); 85\% of stops by 20\% progress & which step broke \\
Trace & latency, delivery status, checklist & per-mode baselines; \textit{D}, \textit{S}, item pass rates & 3.6\% fail; answers within 5 s fail 38.7\% & was this turn delivered \\
Conversation & request replay, context repair, session termination & friction \textit{F}; rule-derived outcome & 5.6\% with failure; 16.7\% at 5+ turns & what users went through \\
Engagement & return, gap, redo, business events & retention net of redo; censoring flag & 10.3\% of users met failure; redo 38.6\% (6.4\% after success) & did the outcome hold \\
\bottomrule
\end{tabular}
\caption{Atomic data, analysis and case performance at the four levels (all real-user data).}
\end{table*}
\textbf{Data and run.} Our case is the business-analysis agent SharpInsight, operated by our team, using its full dialogue from a selected period. The agent and its users operate in Chinese. It offers normal chat, short “simple” answers and asynchronous deep reports. Provenance first removed the 16.1\% of raw turns that were not real traffic: one internal test account and suggested-prompt clicks (whose deep reports were cached, with a non-completion rate of 0.2\% against 7.9\% for real traffic). \textbf{All numbers below use real users only}: 8,537 turns, 4,393 sessions and 1,136 users. Execution and Results data were acquired by rules over all data. Substance, Trajectory and Conduct data were acquired by Claude Opus 5.5 on 107 stratified sessions (267 turns), with Claude Sonnet 5.5 acquiring them independently for adjudication. The sample oversamples hard sessions, and population estimates are corrected with stratum weights \textit{w}\textsubscript{h} = \textit{N}\textsubscript{h}/\textit{n}\textsubscript{h}. All results are aggregates, and no user text is quoted.

\textbf{Span level.} Table~2 summarizes the four levels. Synchronous turns contain 22,884 spans, 3.31 per turn. Logs record step type and order but not per-step time, so the basic data item is step-chain completeness. In normal mode, turns without the thinking step fail 15.7\% of the time against 0.0\% with it; without retrieval, synchronous turns fail 21.6\% against 1.0\%. Of abnormal deep-task stops, 85\% occur at or before 20\% progress, at a median of 60 s against 15~min for completed tasks, so trouble concentrates in task start-up. These chain anomalies accompany failure and are not necessarily its cause. Context reconstruction runs on 91\% of synchronous follow-up turns; among normal-mode turns with retrieval, median latency is 58 s with it against 48 s without.

\textbf{Trace level.} Delivery fails on 3.6\% of turns, highest in deep mode (8.0\%). Latency baselines must be per mode (normal 50 s, simple 8 s, deep 15~min). The most diagnostic indicator is at the other end: normal answers within 5 s fail 38.7\% of the time, against \mbox{0.0–0.2\%} between 10 and 120 s. Substance data are also acquired here.

\textbf{Conversation level.} Sessions with a failure are 5.6\%, rising with length from 3.4\% for one turn to 16.7\% for five or more; sessions with a request replay rise from 16.7\% at two turns to 35.3\% at five or more. By rule, the 107 sessions divide into 14 smooth, 34 struggled, 13 blocked and 46 censored, with mean friction 0.65.

\textbf{Engagement level.} 10.3\% of users met a failure. When a session ends in failure, the next session is a redo in 38.6\% of cases, against 6.4\% after a successful end, with median return gaps of 9 minutes and 18.2 hours. Single-turn sessions are 63.4\% of sessions, yet 71.5\% of them are followed by a return, so delayed labels cover most sessions and only 28.5\% stay censored.

The levels differ clearly. The same failures affect 3.6\% of turns, 5.6\% of sessions and 10.3\% of users: small low-level rates accumulate upward. Each level also sees only its own problems: spans locate which step broke but not whether users suffered, while engagement shows redo but not which step caused it.

\begin{figure*}[t]
\centering
\renewcommand{\thefigure}{3}
\includegraphics[width=\linewidth]{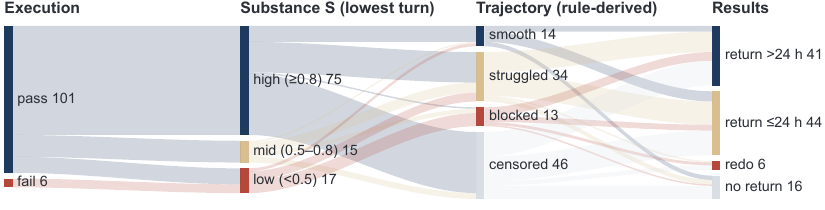}
\caption{The 107 judged sessions along the spine; every stage is computed from atomic data by rule. Trajectory censoring means a single turn shows no process, unlike no return in Results.}
\end{figure*}
\textbf{Substance: item by item.} Substance is not a model’s overall score. Each checklist item answers only yes or no, and the 6 that pass agreement admission enter \textit{S}. With stratum weights, population \textit{S} = 0.92 (95\% CI 0.89–0.95): deep 0.95, normal 0.92, simple 0.84, with no significant deep-normal difference. This differs from the “deep reports score lowest” of LLM overall scores, which conflates form with content. These \textit{S} values use clipped text; the 20 deep reports checked on full text score 0.79, so we do not rank modes by \textit{S}.

Figure~2 breaks this down. Form items are generally high: concrete and actionable are at least 83\% in all three modes. The trouble lies in full-text consistency: no-conflict is about 90\% on clipped text, yet only 1 of 20 reports achieves it on full text, so conflicts surface only on a complete reading. Weak spots also differ by mode: in the judged sample, 11\% of normal answers do not conclude properly, and simple answers follow user constraints only 30\% of the time (\textit{n}~=~10), the latter failing admission and so descriptive only.

\begin{figure}[H]
\centering
\renewcommand{\thefigure}{2}
\includegraphics[width=\linewidth]{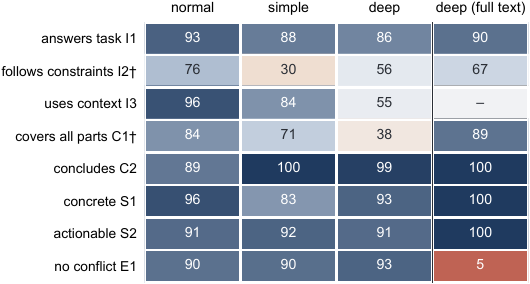}
\caption{Checklist pass rates by mode (\%). The right column shows full-text results for 20 deep reports; the others use clipped text. † Did not pass cross-model admission and is excluded from \textit{S}; “–” marks fewer than 5 applicable cases; values are unweighted.}
\end{figure}
\textbf{Trajectory and Results.} Figure~3 shows the 107 judged sessions along the spine. Failures propagate along it (Figure~4). After a failed turn, the next message is a request replay in 51.5\% of cases, against 12.6\% after a success; the same rule gives 8.6\% on public data, a rough reference since its recall there is only 0.21. Criterion relations point the expected way: mean minimum \textit{S} is 0.32 for blocked sessions, 0.78 for struggled and 0.84 for smooth, and 15\% of blocked sessions are redone with 38\% returning within an hour, against 0\% and 21\% for smooth ones, though with only 13 and 14 sessions these gaps are not significant.

\textbf{Conduct and complementarity.} The system prompt compiles into five clauses. It authorizes disclosing the operating company under in-depth questioning, which we read as meaning the agent should not otherwise volunteer it; by rule count, 134 answers mention the company with no such question. The file-summary template has \textit{C} = 25\% (\textit{n}~=~32) and misfires 93 times with no file uploaded. Of the 107 judged sessions, 55 are flagged by at least one component, yet Execution flags only 6, so trace monitoring alone would miss 89\% of problem sessions. Trajectory, Conduct and Substance carry 27, 6 and 1 unique flags, and pairwise Jaccard overlap ranges from 0.00 to 0.35. Of the 101 sessions that pass Execution, 26 still have a minimum \textit{S} below 0.8 (Figure~3).

\section{How the Loop Runs}
These results are not the product of one pass but of the loop’s successive corrections.

\begin{table*}[t]
\centering
\renewcommand{\thetable}{3}
\fontsize{8}{10}\selectfont\setlength{\tabcolsep}{4pt}
\begin{tabular}{@{}>{\raggedright\arraybackslash}p{\dimexpr 0.140\linewidth-8pt\relax}>{\raggedright\arraybackslash}p{\dimexpr 0.380\linewidth-8pt\relax}>{\raggedright\arraybackslash}p{\dimexpr 0.250\linewidth-8pt\relax}>{\raggedright\arraybackslash}p{\dimexpr 0.230\linewidth-8pt\relax}@{}}
\toprule
\textbf{Audit} & \textbf{Finding} & \textbf{Inner-loop action} & \textbf{After correction} \\ \midrule
Provenance check & 13 of 120 sampled sessions came from one internal account & move cleaning before sampling & 107 sessions, all real users \\
Evidence coverage & a median of about 6\% of deep-report text reached evaluators & widen context projection to full text & same reports: \textit{S} 0.92 → 0.79 \\
Judgment method & direct LLM scores have large residuals; outcome κ only 0.63 & switch to atomic data plus deterministic analysis & ICC 0.81 → 0.86; κ 0.63 → 0.81 \\
Rule lexicon & 87\% of “human” (\raisebox{-.12em}{\includegraphics[height=.8em]{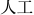}}) rule hits were “AI” (\raisebox{-.12em}{\includegraphics[height=.8em]{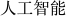}}) & replace phrases; add false-friend list & 0 hits in the full log after correction \\
Goodhart test & raw 7-day return higher after failure (79.1\% vs. 60.4\%) & redefine net of redo & no longer higher after failure (47.5\% vs. 55.7\%, intervals overlap) \\
Agreement between models & low-agreement items counted as flags, e.g. overclaiming 53 by Opus, 15 by Sonnet (κ 0.21) & remove pending a new definition & Conduct flags 63 → 14 \\
Conclusion stability & “deep reports lowest” and “report generation resolves least” varied with method or model & withdraw; restate item by item & complete in form, inconsistent on full text \\
\bottomrule
\end{tabular}
\caption{Inner-loop corrections that actually occurred in the case.}
\end{table*}
\textbf{Inner loop: evaluation corrects itself.} The audit applies the measurement model to data. Treating Opus and Sonnet as two raters on the 259 delivered turns, 85.8\% of the variance in \textit{S} comes from the responses, systematic rater bias is about zero and the residual is 14.2\%, giving a single-rater ICC of 0.86 and a two-rater jury of 0.92. By contrast, direct 1–5 LLM scoring on the same turns has a residual of 19.0\% and an ICC of 0.81. Trajectory outcomes tell the same story: rule-derived from atomic data, cross-model κ = 0.81; as direct LLM labels, only 0.63. This is the value of separating acquisition from analysis.

\begin{figure}[H]
\centering
\renewcommand{\thefigure}{4}
\includegraphics[width=\linewidth]{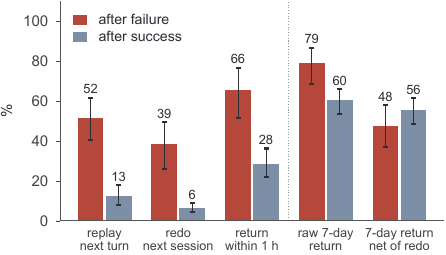}
\caption{User behavior after delivery failure vs. success (all real-user data; whiskers show user-clustered 95\% CIs). Left of the dotted line is the failure chain; right of it is retention. Redo and one-hour return use sessions with a next session as the base.}
\end{figure}
The data admission line is κ \ensuremath{\geq} 0.60 at two decimals (Figure~5); session termination and context repair are 0.599 and 0.596 unrounded. Verbatim replay (0.93), concludes (0.93) and the disclosure clause (1.00) are reliable; overclaiming (0.21), the case clause (0.30) and scope mismatch (0.38) change with the model and leave the analysis. The audit also quantifies the recall gap of rules: against the atomic data, Execution rules have precision 1.00 and recall 0.33, and the replay rule 0.94 and 0.46, which determines what LLMs must acquire.

Table~3 lists seven inner-loop corrections that actually occurred, spanning evidence, judgment method, rules, metrics and conclusions. Evidence scope is the clearest. To control cost, the first pass kept only the head and tail of long answers, and the audit found that about 6\% of a deep report’s text reached evaluators. On full text, \textit{S} for the same 20 reports fell from 0.92 to 0.79, and the no-conflict item fell from 90\% to 5\%.

\textbf{Outer loop: per-component regression.} The outer loop requires a new version to pass regression checks on every component. Before the gate meets a real release, two questions need answers: does it raise false alarms when nothing changed, and how large a regression can it catch? Two offline experiments test the gate itself. Both randomize by user and avoid any time split (Table~4).

\textbf{A/A test.} We split all users at random into two halves 4,000 times (Kohavi, Tang, and Xu 2020). The halves have no real difference, so any verdict is a false alarm. A per-turn test (|\textit{z}| \ensuremath{\geq} 2) raises at least one false alarm in 98\% of splits, because turns from one user are not independent and a few heavy users can move a rate. The gate therefore uses a user-clustered test, with per-metric thresholds calibrated on half the splits. On the held-out half, the seven metrics together false-alarm 4.5\% of the time, within the 5\% target.

\textbf{Seeded defects.} On one half of the users we rewrite the logged turns (blanking answers, resending the previous request, inserting a disclosure sentence or template) and recompute with the same rules; latency is scaled up. Each defect runs 300 times at each strength. At 80\% detection, the gate catches a rise of 3.2 points in failure, 2.1 in disclosure and 1.5 in template misfires, and a 46\% latency increase, but it is insensitive on replay (18.1 points) and replay after failure (35.5 points). Defects rarely spill across components: Trajectory and Conduct defects flag other components at most 2\% of the time. The exception is that injected failures change the denominator of replay after failure, which is flagged in about 16\% of runs at 95\% detection strength. So false alarms are controlled, yet at current traffic the gate catches only large regressions. These experiments test the gate, not any particular change, and Substance is left out because it depends on sampled audits.

\begin{figure}[H]
\centering
\renewcommand{\thefigure}{5}
\includegraphics[width=\linewidth]{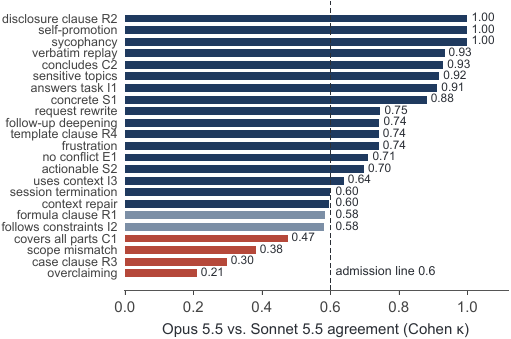}
\caption{Data admission. Agreement between two models per data item (to two decimals); blue enters deterministic analysis; grey and red fall below the line and await new definitions.}
\end{figure}
\textbf{From findings to fixes.} Each finding becomes a fix bound to an acceptance metric. The case yields three: clarify very short inputs before launching a report, accepted on the failure rate of 1–5 character prompts (7.3\%); tighten template triggers, accepted on misfire rate and compliance (25\%); retry empty and truncated generations, accepted on failure within 5 s (38.7\%) and redo after failure (38.6\%). These fixes are not yet live; once live they will pass through the same gate.

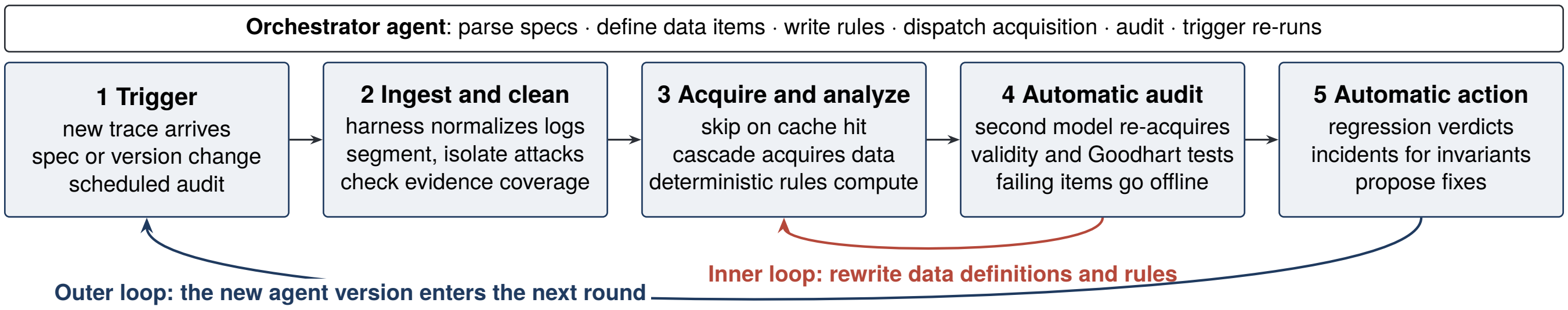
\begin{figure*}[t]\centering
\renewcommand{\thefigure}{6}
\resizebox{\linewidth}{!}{\begin{tikzpicture}[x=1pt,y=1pt,font=\sffamily\fontsize{7.2}{9}\selectfont,
 bx/.style={draw=navy,fill=mist,line width=.5pt,rounded corners=1.5pt,minimum width=92pt,minimum height=50pt,align=center,inner sep=1pt},
 arr/.style={-{Stealth[length=4pt,width=3pt]},line width=.6pt,color=ink}]
\node[draw=ink,line width=.5pt,rounded corners=1.5pt,minimum width=505pt,minimum height=15pt,inner sep=1pt] at (252.5,66) {\textbf{Orchestrator agent}: parse specs · define data items · write rules · dispatch acquisition · audit · trigger re-runs};
\node[bx] (a) at (46,30) {\textbf{\fontsize{7.8}{10}\selectfont 1 Trigger}\\[1pt]new trace arrives\\spec or version change\\scheduled audit};
\node[bx] (b) at (149.25,30) {\textbf{\fontsize{7.8}{10}\selectfont 2 Ingest and clean}\\[1pt]harness normalizes logs\\segment, isolate attacks\\check evidence coverage};
\node[bx] (c) at (252.5,30) {\textbf{\fontsize{7.8}{10}\selectfont 3 Acquire and analyze}\\[1pt]skip on cache hit\\cascade acquires data\\deterministic rules compute};
\node[bx] (d) at (355.75,30) {\textbf{\fontsize{7.8}{10}\selectfont 4 Automatic audit}\\[1pt]second model re-acquires\\validity and Goodhart tests\\failing items go offline};
\node[bx] (e) at (459,30) {\textbf{\fontsize{7.8}{10}\selectfont 5 Automatic action}\\[1pt]regression verdicts\\incidents for invariants\\propose fixes};
\draw[arr] (a)--(b); \draw[arr] (b)--(c); \draw[arr] (c)--(d); \draw[arr] (d)--(e);
\draw[-{Stealth[length=5pt,width=4pt]},line width=.8pt,color=acc] (d.south) .. controls (355.75,-8) and (252.5,-8) .. (c.south);
\node[text=acc,anchor=north] at (304,-7) {\textbf{Inner loop: rewrite data definitions and rules}};
\draw[-{Stealth[length=5pt,width=4pt]},line width=.8pt,color=navy] (e.south) .. controls (459,-30) and (46,-30) .. (a.south);
\node[fill=white,inner sep=1.5pt,text=navy] at (112,-20) {\textbf{Outer loop: the new agent version enters the next round}};
\end{tikzpicture}}
\caption{The automation pipeline for closed-loop evaluation. The orchestrator agent links five steps; the red and blue arrows are the inner and outer loops.}
\end{figure*}
\section{Porting to Other Agents}
The method’s generality comes from its layering. The measurement model, levels, components, loops and admission rules stay fixed, and a port replaces three things: the \textbf{specification} sets clauses, the \textbf{knowledge pack} sets baselines, lexicons and data definitions, and the \textbf{harness} sets which logs and outcome events connect. The deterministic analysis \textit{f} mostly carries over. Take Substance and Results: a customer-service agent would check whether replies cite policy and issues get resolved; a coding agent, whether tests pass and code is not reverted; a data-analysis agent, whether definitions are stated and reports are adopted. These mappings are reasoned, not yet tested one by one.

We also ran a test on public data, applying the Trajectory acquisition prompt unchanged to 60 Chinese WildChat conversations (164 turns; Zhao et al. 2024) from a general-purpose assistant, a different domain and user base. Acquisition ran without modification, and the same rules gave 17 smooth, 17 struggled, 3 blocked and 23 censored, whereas direct LLM labels left 50 unknown. Keyword rules did not transfer: replay rule precision fell from 0.94 to 0.75 and recall from 0.46 to 0.21. This marks the boundary of porting: \textbf{atomic data acquisition and deterministic analysis transfer, keyword rules must pass the audit again, and the Results stage must connect to new event sources.}

\section{Making the Loop Run Automatically}
Closed-loop evaluation only makes sense when automated. People can run one audit; they cannot do it continuously for every turn and version. The key is an \textbf{orchestrator agent} that organizes the Agent Factory’s whole pipeline (Figure~6).

\textbf{Five steps.} (1) \textbf{Trigger}: a new trace starts incremental evaluation, a specification or version change starts recomputation, and a schedule starts the audit. (2) \textbf{Ingest and clean}: the harness normalizes logs, and Provenance segments traffic and checks evidence coverage. (3) \textbf{Acquire and analyze}: on a cache miss, rules or LLMs acquire data through the cascade in parallel, and deterministic rules compute each level’s results. (4) \textbf{Automatic audit}: a sample goes to a second model for re-acquisition, agreement and validity are computed, and failing data items leave the analysis and trigger a definition rewrite. (5) \textbf{Automatic action}: new versions are compared component by component, invariant breaches open incidents, and frequent defects become proposed fixes.

\textbf{How the factory builds evaluators.} The orchestrator reads the role specification and system prompt, generates data definitions, evidence scopes and output schemas, writes deterministic rules as versioned code, and rewrites them when the audit fails. People define the component split and invariants. When a specification changes, clauses and data definitions update, affected results are recomputed, and the rest stay in cache.

\textbf{Measured run.} The case study was produced this way. After a person set the component split, the orchestrator wrote the rules, defined data items, dispatched batches, ran the statistics and performed the audit. It located and carried out the corrections in Table~3, while people set requirements and reviewed results. Evaluation used 19 agents across two parallel rounds, about 2–7 minutes per agent and about 4M tokens cumulatively excluding cache reads, with no per-item human intervention. A log replay indicates that incremental evaluation cuts turn-evaluations from 28,044 to 8,537 (\ensuremath{-}70\%).

\begin{table}[H]
\centering
\renewcommand{\thetable}{4}
\fontsize{8}{10}\selectfont\setlength{\tabcolsep}{4pt}
\begin{tabular*}{\linewidth}{@{\extracolsep{\fill}}lrrrr@{}}
\toprule
\textbf{Metric} & \textbf{Base} & \textbf{Per-turn} & \textbf{Calibrated} & \textbf{Min. rise} \\ \midrule
Failure & 3.6\% & 36\% & 0.9\% & +3.2 pp \\
Deep failure & 8.0\% & 30\% & 0.3\% & +9.7 pp \\
Normal latency & 40 s & 63\% & 0.9\% & +46\% \\
Replay & 13.6\% & 76\% & 0.8\% & +18.1 pp \\
Replay after failure & 51.5\% & 18\% & 0.6\% & +35.5 pp \\
Disclosure & 1.6\% & 34\% & 0.7\% & +2.1 pp \\
Template misfire & 1.1\% & 24\% & 0.5\% & +1.5 pp \\
\textbf{Any metric} &  & \textbf{98\%} & \textbf{4.5\%} &  \\
\bottomrule
\end{tabular*}
\caption{Offline test of the gate (all real-user data). Per-turn and Calibrated are A/A false-alarm shares, the latter on held-out splits; Min. rise is the increase above baseline that reaches 80\% detection. Latency base is the geometric mean.}
\end{table}
\textbf{Where people stay.} People keep three decisions: defining invariants, approving specification changes and reviewing audit samples. The current implementation is still offline batch processing. Online triggers, gold sets and the release gate are not yet connected to production, and outer-loop fixes have not been validated through a real release.

\section{Discussion}
\textbf{What the method changes.} Closed-loop evaluation turns evaluation from a one-off score into a running control system, and separating acquisition from analysis keeps that system explainable and auditable. The case’s most serious problems sit not in infrastructure but in job design: one or two characters can launch a report of over ten thousand characters, and deep mode rewards length over consistency. Such problems surface only through layered measurement and are fixed only through the outer loop. Just as important, the method can overrule itself: metrics were retired and conclusions withdrawn, which open-loop evaluation cannot do.

\textbf{Limitations.} The case covers one agent and one language, and the judged sample is small. Checklists mostly used clipped text, and full-text checks cover only 20 deep reports. Form items in \textit{S} nearly saturate, limiting its discrimination, and some data items still need better agreement. Agreement between models does not mean correctness, so a human gold set is the next step, and the κ line is an empirical choice. Retention net of redo removes exactly the returns more common after failure, so part of the reversal comes from the definition, and the two intervals overlap. Execution rules have recall of only 0.33, so rule-based failure rates may be understated. Seeded defects assume random occurrence across users, while real defects may cluster in some scenarios. Our contrasts are observational, and causality needs controlled experiments (Kohavi, Tang, and Xu 2020; Bickel, Hammel, and O’Connell 1975).

\textbf{Relation to prior work.} This method complements capability benchmarks (Yao et al. 2025; Jimenez et al. 2024; Xu et al. 2025) and their methodological critiques (Kapoor et al. 2025, 2026; Yehudai et al. 2026). It carries LLM and agent judges (Zheng et al. 2023; Wang et al. 2024; Zhuge et al. 2025; Shankar et al. 2024) and failure diagnosis (Cemri et al. 2025; Zhang et al. 2025) into production, and it draws on implicit-satisfaction research (Hassan et al. 2013; Lin et al. 2024; Rodden, Hutchinson, and Fu 2010), atomic fact checking (Min et al. 2023) and ML-ops practice (Sculley et al. 2015). Conduct draws on alignment research (Bai et al. 2022; Sharma et al. 2024) and the distinction between task and contextual performance (Borman and Motowidlo 1993).

\section{Conclusion}
Evaluating an agent, like reviewing an employee, takes more than one judge and one score. We proposed closed-loop evaluation: every level’s conclusion is computed by rules from traceable evidence, an inner loop calibrates evaluation itself, an outer loop drives agent improvement, and an orchestrator agent links these steps into an automated pipeline. The method’s levels, components and loops are not tied to a specific domain. We tested its core components on SharpInsight, checked the portability of some on public conversation data, and tested the outer-loop gate offline; the real effect of the fixes and the method’s performance on other agent types await real releases and further study. Evaluation is finished only when it can correct itself and change the agent. That is what closing the loop means.

\balance
\makeatletter\renewcommand\@biblabel[1]{}\makeatother
\renewcommand\refname{References}
{\fontsize{8}{9.6}\selectfont\raggedright
}
\end{document}